\documentclass[conference]{IEEEtran}
\IEEEoverridecommandlockouts
\usepackage{cite}
\usepackage{amsmath,amssymb,amsfonts}
\usepackage{algorithmic}
\usepackage{graphicx}
\usepackage{textcomp}
\usepackage{xcolor}
\usepackage{todonotes}
\usepackage{tabularx,booktabs}
\def\BibTeX{{\rm B\kern-.05em{\sc i\kern-.025em b}\kern-.08em
    T\kern-.1667em\lower.7ex\hbox{E}\kern-.125emX}}
\begin{document}

\title{On the Definition of Appropriate Trust\\ and the Tools that Come with it.\\

\thanks{Helena Löfström is a PhD student in the Industrial Graduate School in Digital Retailing (INSiDR) at the University of Borås, funded by the Swedish Knowledge Foundation, grant no. 20160035.}
}

\author{\IEEEauthorblockN{Helena Löfström}
\IEEEauthorblockA{\textit{Jönköping International Business School, Jönköping University, Sweden} \\ %dept. name of organization (of Aff.)
\textit{Department of Information Technology, University of Borås, Sweden}\\%name of organization (of Aff.)
%Allégatan 1\\ 50190
 % Borås\\
helena.lofstrom@ju.se, 0000-0001-9633-0423}%email address or ORCID
}

\maketitle

\begin{abstract}
%This document is a model and instructions for \LaTeX.
%This and the IEEEtran.cls file define the components of your paper [title, text, heads, etc.]. *CRITICAL: Do Not Use Symbols, Special Characters, Footnotes, or Math in Paper Title or Abstract.
% MAX 8 SIDOR INKLUSINVE REFERENSER
Evaluating the efficiency of human-AI interactions is challenging, including subjective and objective quality aspects. With the focus on the human experience of the explanations, evaluations of explanation methods have become mostly subjective, making comparative evaluations almost impossible and highly linked to the individual user. However, it is commonly agreed that one aspect of explanation quality is how effectively the user can detect if the predictions are trustworthy and correct, i.e., if the explanations can increase the user's appropriate trust in the model. This paper starts with the definitions of appropriate trust from the literature. It compares the definitions with model performance evaluation, showing the strong similarities between appropriate trust and model performance evaluation. The paper's main contribution is a novel approach to evaluating appropriate trust by taking advantage of the likenesses between definitions. The paper offers several straightforward evaluation methods for different aspects of user performance, including suggesting a method for measuring uncertainty and appropriate trust in regression.
\end{abstract}


\section{Introduction} \label{chap:intro}
%\todo[inline]{Hur utvärderar man förklaringar, de metriker som finns, prooblemet med att det är omöjligt med a obejktiv utvärdering av mänskliga användare om vi använder de subjektiva metrikerna. Vi behöver använda oss av objektiva metriker, vilka indikerar hur väl förklaringarna hjälper användarna att identifiera vilka prediktioner som är bra och vilka som är dåliga}

Artificial intelligence (AI) has become a universal component of decision support systems (DSSs) across various domains, such as retail or defence \cite{zhou2021evaluating}. These AI-based predictive models are generally very accurate but are often opaque and lack transparency and comprehensibility, making it difficult for users to understand how predictions are generated. Furthermore, there is a risk that models may become biased based on something in the data, leading to erroneous predictions. Thus, it is crucial that these models are designed to provide explanations of their predictions, enabling users to detect and correct potential errors. The lack of transparency in DSSs can otherwise lead to low acceptance and mistrust of such systems \cite{ribeiro2016should}. The risks of a lack of transparency have been highlighted in regulations from numerous government initiatives, such as the European Guidelines for Trustworthy AI \cite{AIHLEG}, which emphasise the need for accountability, transparency, and human oversight in AI systems.

Explainable artificial intelligence (XAI) is a field dedicated to creating AI systems that can explain their rationale to human users. The goal of XAI is to provide explanations that characterise the strengths and weaknesses of the underlying model and communicate how it will behave in the future \cite{DavidGunning2017,dimanov2020you}. However, evaluating the quality of explanations can be challenging, as it involves the interaction between humans and machines, requiring multiple aspects of quality to be considered \cite{Lofstrom2022}. 

The motivation for this work is twofold. Firstly, the area is in dire need of rigour for definitions and benchmarks of objective metrics to evaluate the interaction between humans and computers in explanations. There are a plethora of subjective measurements to use \cite{hoffman2018metrics}, making it challenging to conduct comparative evaluations. One metric; called \textit{appropriate trust}, has recently risen as a promising solution to the problem by measuring user reliance, resulting from the subjective criteria. Although the meaning behind appropriate trust is generally understood as accurate trust in the predictions, the researchers are still left to find their own methodology for evaluating different aspects of the metric since formalised definitions of the method are almost absent \cite{Guidotti18}. Secondly, by defining straightforward and easy-to-follow methods of evaluating various elements of appropriate trust, the rigour in the area is strengthened, and comparative evaluations become possible. 

The main contributions of this paper are:
\begin{itemize}
    \item The proposal of a well-formulated method for evaluating appropriate trust, which can facilitate more accurate and consistent assessments of user performance.
    \item To address the challenge of evaluating user interaction with explanation methods, the paper proposes well-defined methods for measuring misuse or disuse in a user evaluation.
    \item In addition, the paper also suggests a method for measuring uncertainty in regression during user evaluations.
\end{itemize}

% ====================================================
% STRUCTURE OF PAPER
% ====================================================
The rest of the paper is organised as follows: the next section reviews fundamental concepts related to explanation methods. Moreover, it provides an introduction to evaluation of model performance. Section~\ref{definitions} presents earlier definitions and usage of appropriate trust, while Section~\ref{user_perf} defines methods for measuring misuse and disuse. Section~\ref{uncert} introduces uncertainty estimation for regression. In Section~\ref{User_ev} the metrics are exemplified with the results from a user evaluation. The paper ends with a discussion, followed by concluding remarks.
% ==================================================
\section{Background}
\subsection{Post-Hoc Explanations}
In machine learning, generating explanations can be approached in two ways: developing inherently interpretable and transparent models or using \textit{post-hoc} methods to explain complex models. Post-hoc explanations involve creating more straightforward and interpretable models that explain how the complex model's prediction relates to input features. These explanations can be either local (explaining a single instance) or global (explaining the model) and often include visualisations, such as feature importance plots, pixel representations, or word clouds, to highlight the most important features, pixels, or words that drive the model's predictions \cite{molnar2020interpretable, moradi2021post, holzinger2019causability}.

Local explanations in classification can be based on probability estimates, which most machine learning (ML) models can provide as an indication of the likelihood of each class. These probability estimates are commonly interpreted as a measure of prediction quality. For instance, in a binary classification problem, a model predicting an instance belonging to the positive class with a probability estimate of 0.89 would be considered more certain than a model that predicts the same instance with a probability estimate of 0.69. Probability estimates can thus serve as a foundation for local explanations in classification tasks.

\subsection{Evaluation of Explanations}
Explanations in AI exist in an interaction between humans and computers; consequently, objective and subjective metrics are necessary to evaluate the success of the explanations. While objective measurements provide quantitative insights into the effectiveness of the explanations, subjective measurements are also critical for the system's success, since they reflect the user's perception and experience \cite{DavidGunning2017}. There is an ongoing discussion in the field about how to evaluate explanations best, and so far, there is no agreed-upon metric of how to measure quality \cite{zhou2021evaluating}. However, one goal for explanations often referred to in the literature is to increase the user's trust in the system \cite{doshi2017towards, gunning19, Carvalho19, hoff2015trust,zhou2021evaluating}. However, the subjective nature of the trust makes evaluations challenging, urging some authors to even suggest that comparative evaluations are impossible \cite{zhou2021evaluating}. One metric has in recent years risen as a promising solution to the problem; \textit{appropriate trust}\cite{gunning19} (also called \textit{calibrated trust}\cite{jacovi2021formalizing} or \textit{justified trust}\cite{hoffman2018metrics}). Appropriate trust is being understood as to what extent the user can detect the correct and incorrect predictions, or in other words, if the users trust the correct predictions and mistrust the incorrect \cite{yang20}. 

\subsection{Misuse and Disuse}
It is not enough to only measure how much the users trust the system since they may trust both correct and erroneous predictions. When the user has a low level of appropriate trust in the system, two unwanted situations could occur, \textit{misuse} or \textit{disuse}. Misuse and disuse are two situations that explanations are meant to avoid and which becomes possible to detect when measuring the level of appropriate trust. However, there is no clear and commonly agreed notion of how to measure the level of the terms.

\subsubsection{Misuse} is a tendency of the user to have an over-reliance on the system, which can result in users failing to identify failures or decision biases \cite{sheridan2002humans,adya2020stressed}. It can also be defined as a higher trust than appropriate \cite{alvarado2014reliance,buccinca2020proxy}. With a higher than appropriate reliance on the system, the user put too high trust in the system, neglecting possible errors or doubts. In that situation, an evaluation of the user's performance would result in a high number of predictions trusted as correct, including a too-high number of incorrect predictions. The term is also called \textit{overtrust} in \cite{yang20}.

\subsubsection{Disuse} is often identified as more problematic since it refers to reliance in a system that is lower than appropriate, given the actual performance of the system \cite{alvarado2014reliance,buccinca2020proxy}. In extreme situations, it could lead to neglect or under-utilisation of the system \cite{alvarado2014reliance}. The term is also called \textit{undertrust} in \cite{yang20}.

\subsection{Metrics for Model Performance in Classification}
An effective and often-used strategy for measuring model performance in classification is to look at the confusion matrix. The general idea is to count the number of times the model confuses the predictions, i.e., attach the wrong label to the instances. The number of correctly and incorrectly predicted instances are then compared, and several different aspects of model performance can be calculated \cite{geron2022hands}. Two of the most commonly used metrics are \textit{precision} and \textit{recall}.

\begin{figure}
    \centering
    \includegraphics[width= 0.8\linewidth]{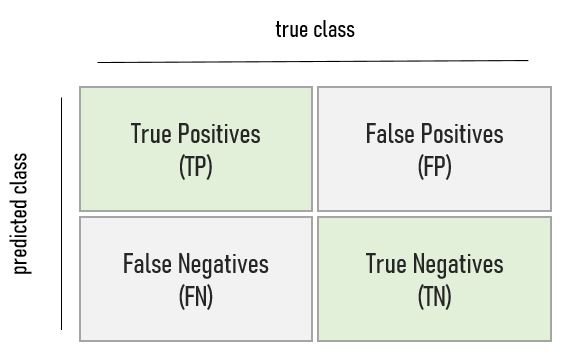}
    \caption{Confusion matrix for model performance in Classification.}
    \label{fig:my_label}
\end{figure}

\subsubsection{Precision}
is also called the proportion of the correct positive predictions. A high precision means that the model has a high number of true positives together with a limited number of false positives. In other words, the model succeeds at correctly predicting the instances \cite{geron2022hands}. The metric is written as the number of true positives divided by the total number of predicted positives, as follows:
\begin{equation*}
    precision = \frac{TP} {TP + FP} 
\end{equation*}

\subsubsection{Recall}
also called sensitivity, is typically used together with precision. The metric is used to identify the proportion of positive instances correctly identified by the model. A high recall signals that the model can successfully identify the correct instances without mislabelling them \cite{geron2022hands}, and is calculated as follows:

\begin{equation*}
    recall = \frac{TP} {TP + FN} 
\end{equation*}

\subsubsection{$F_1$ score}
Precision and recall can be combined into a single metric called the $F_1$~score. The metric calculates the harmonic mean between recall and precision and gives much more weight to low values, with the result that the classifier only gets a high $F_1$~score if both recall and precision are high \cite{geron2022hands}. The metric is calculated as follows:

\begin{equation*} 
    F_1 = 2 * \frac{precision * recall}{precision + recall}
\end{equation*} 

\subsection{Conformal Regression and prediction intervals}
Conformal Regression generates valid prediction intervals that quantify the uncertainty of each individual prediction \cite{vovk2005algorithmic}. The method is applied on top of any machine learning model and transforms the model's point predictions into intervals that are guaranteed to include the true target with the confidence of $1-\epsilon$ (a user-assigned significance level). This is a valuable element that sets Conformal Regression apart from other learning frameworks, as it allows you to assess your confidence level in each prediction. Through the use of prediction intervals, Conformal Regression can provide bounds specific to each instance \cite{johansson2014regression} rather than just providing a single bound for the entire distribution. This is particularly useful in practical applications where specific predictions require a higher level of certainty than others. In summary, Conformal Regression is a powerful tool for making accurate predictions while also providing a measure of uncertainty through prediction intervals. This can notably enhance the reliability and usefulness of machine learning models in a wide range of real-world applications.

\subsection{Venn-Abers and uncertainty estimation}
Venn-Abers predictors \cite{Lambrou2015} are used to generate probability estimate intervals for binary classification that incorporates the uncertainty of each class. The method is applied, just as Conformal regression, on top of any machine learning model and generates a well-calibrated probability interval for each class. The size of the interval indicates the confidence associated with the probability interval, with a smaller interval indicating higher confidence and a wider interval indicating lower confidence. For a model to be well-calibrated, the predicted probabilities must be matched by observed accuracy. This means that predictions with a probability estimate of $0.85$ should be correct in $85\%$ of the cases.

\section{Definitions of Appropriate Trust} \label{definitions}
In this section, earlier definitions of appropriate trust are presented. The definitions are then discussed in light of how model accuracy is evaluated in \textit{machine learning} (ML). It is worth noticing that this paper's presented sample of definitions is mainly chosen from the meta survey in\cite{Lofstrom2022}, covering between (due to presumed overlap) 739 - 2 750 articles. The articles included in this paper were chosen based on their definitions of appropriate trust. 

Frequent references in the literature are made to evaluations of whether the user is appropriately trusting the system \cite{mcdermott2019practical,Carvalho19,arrieta20,chromik2020taxonomy,hoff2015trust,doshi2017towards,das2020opportunities,adadi2018peeking,hoffman2018metrics,wang2019Designing,mohseni2018multidisciplinary,gunning2021darpa,zhang2018explainable}. In this paper, the name appropriate trust has been chosen to create a coherent understanding of the term, although, in the cited papers, the authors may use different names. In some papers \cite{mohseni2018multidisciplinary}, the metric is called simply trust, while others justified trust \cite{hoffman2018metrics}, calibrated trust \cite{jacovi2021formalizing}, perceptual accuracy \cite{merritt2015well}, or causability \cite{holzinger2019causability}.

\subsection{Previous Definitions}
\textbf{John D. Lee and Katrina A. See (2004)}. 
In \cite{lee2004trust}, the authors define inappropriate reliance as associated with misuse and disuse and write that it partly depends on how well the trust matches the true capabilities of the system.

\textbf{S. M. Merritt et al. (2015)}. 
In \cite{merritt2015well}, the authors write that appropriate reliance should be reached and misuse and disuse avoided to achieve maximum safety and performance. Appropriate trust is later interpreted as the correspondence between the user's trust in the system and the capabilities of the system. The author further states that greater correspondence between actual and perceived reliability at a given time reflects a higher trust calibration. It can thus be measured by assessing a user's beliefs about aid reliability and comparing those beliefs to the aid's actual reliability level.

\textbf{Hoffman et al. (2018)}. 
In \cite{hoffman2018metrics}, the authors write that the system should help the user to know whether, when, and why to trust and rely upon the system and understand whether, when, or why to mistrust the system. The system should also indicate to the user to rely on or only rely on the system with caution. The authors also point out that users always have a variety of mixing between justified and unjustified mistrust in the system, which might change over time. In \cite{hoffman2018metrics}, the authors define the metric as the output from the subjective measurements and possible to measure with an evaluation where the user indicates if the prediction is trustworthy or not.

\textbf{D. Gunning et. al. (2019)}. 
Appropriate trust is mentioned immediately together with effectively managing and understanding the system in \cite{gunning2021darpa}, and more specifically defined in \cite{gunning2019xai} as an objective measure for an explanation's effectiveness. The authors also point out that the metric should answer questions of the explanation that improves the user's decision-making.

\textbf{A. Holzinger et. al. (2019)}. 
In \cite{holzinger2019causability}, the authors define appropriate trust as "the extent to which an explanation of a statement to a human expert achieves a specified level of causal understanding with effectiveness, efficiency and satisfaction in a specified context to use." The authors later developed in \cite{holzinger2020measuring}, a System Causability Scale metric based on this definition, using Likert Scales and subjective evaluation of explanation effectiveness.

\begin{figure}[h!]
    \centering
    \includegraphics[width = 0.7 \linewidth]{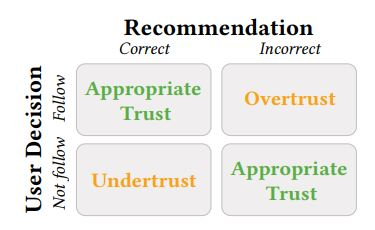}
    \caption{Identifying appropriate trust, from \cite{yang20}.}
    \label{fig:AT_conf_matrix}
\end{figure}

\textbf{F. Yang et. a. (2020)}. 
\cite{yang20} defines appropriate trust based on as \textit{"to [not] follow a [in]correct recommendation"}. The authors further writes that based on the definition, appropriate trust is the alignment between the perceived and actual performance of the system and related to the user's abilities to recognise when the system is correct and when it is incorrect. They use a type of confusion matrix as shown in Fig.\ref{fig:AT_conf_matrix} to visualise how to identify appropriate trust, overtrust, and undertrust.

\textbf{A. Jacovi et. al (2021)}. 
In \cite{jacovi2021formalizing} the authors define appropriate trust as when the trust is calibrated with trustworthiness. The author also points out that if misaligned, it can lead to disuse, misuse and abuse, and that if trust exceeds trustworthiness it leads to misuse, while trustworthiness exceeding trust lead to disuse.

\textbf{Conclusion:} From the definitions, a picture emerges pointing out that the level of appropriate trust should indicate the level of effectiveness of the explanation method in making the user take better decisions. The metric is an alignment between perceived correctness and actual correctness, as is visualised using a confusion matrix in Fig.~\ref{fig:AT_conf_matrix} and is an objective outcome of the subjective measurements closely connected to the user. Notably, the definitions of appropriate trust are based on a classification context, where the prediction is either correctly classified or not.

\subsection{Earlier Usage and Definition of Method}
Although the definition of appropriate trust is straightforward, finding an evaluation with a transparent methodology is not easy. With the descriptions moving from a binary situation of either having or not having appropriate trust to a scale of more or less appropriate trust, the methodology also changes. Some authors emphasise subjective aspects of trust. In, e.g., \cite{hoffman2018metrics}, the authors focus on the user's evaluation of the experience of the system. Whereas others, like \cite{yang20}, are interested in to which degree the users can distinguish erroneous predictions. %Since this paper focuses on finding methods for measuring the outcome of the subjective mental model, the user's evaluation of experience has been excluded.

In the definition from Fig.~\ref{fig:AT_conf_matrix}, in \cite{yang20}, trust is divided into three categories: appropriate trust, overtrust, and undertrust. %In the paper, the authors add another aspect of trust, self-confidence. Since this paper focuses on the objective outcome of the user, the subjective measurement was excluded. 
Appropriate trust is measured by counting how many times the users follow a correct prediction and do not follow an incorrect prediction \cite{yang20}. Overtrust is measured as how often a user follows an incorrect prediction, and undertrust is how often a user does not follow a correct prediction. The three metrics are calculated in the range of $[0,1]$. The derivation of metrics in this paper is partly built on the confusion matrix in \cite{yang20}.

\textbf{Conclusion}
Where there is a transparent methodology, appropriate trust is measured as the users performance. It is answering the questions of how well the explanations can help the user detect the correct and incorrect predictions and, in some way treats the user as a black box; the user is given predictions which are processed with a complex mix of subjective characteristics like expectations or curiosity, resulting in a decision; to trust or not trust the prediction. 
%
% =========================================================================================================================================
%  RESULTS
% =========================================================================================================================================
\section{Measuring Appropriate Trust as User Performance} \label{user_perf}

From earlier sections in the paper, a number of characteristics when measuring appropriate trust were identified:
\begin{itemize}
    \item Appropriate trust could be defined as the \textit{performance of the user}
    %\item when measuring the performance of the user, the \textit{confusion matrix} could be used.
    \item overtrust (or misuse) is defined and measured as when the user places \textit{trust in predictions that should not be trusted}.% instead of the predictions that should be trusted.
    \item undertrust (or disuse) is defined and measured as when the user \textit{does not trust predictions that should be trusted}.% instead of those that should be trusted.
    %\item Total appropriate trust can be measured as if the user can correctly distinguish when to trust or not trust all the given predictions.
    \item Appropriate trust is binary for each prediction; either the user trusts or distrusts the prediction. 
    \item The metric (averaged over many predictions) \textit{vary depending on to which degree the user misplaces trust in the given predictions}. 
\end{itemize}

\begin{figure}[h!]
    \centering
    \includegraphics[width=0.7\linewidth]{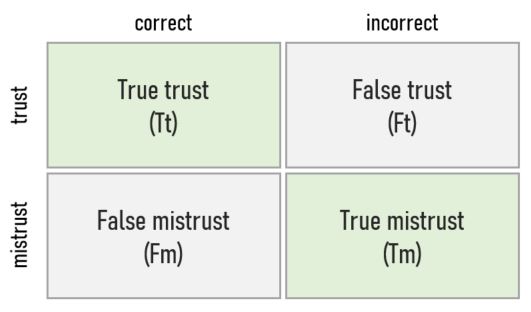}
    \caption{Confusion trust matrix for user evaluations, inspired by \cite{yang20} }
    \label{fig:trust_conf_matr}
\end{figure}

Consequently, when the user is treated as a classifier in an evaluation, identifying predictions as either correct or incorrect, the result is possible to analyse %as in \cite{yang20} 
with a confusion matrix similar to Fig.~\ref{fig:AT_conf_matrix} and Fig.~\ref{fig:my_label}. Fig.~\ref{fig:trust_conf_matr} shows: 
\begin{itemize}
    \item the number of correct predictions correctly trusted, which is called \textit{True trust} ($Tt$); 
    \item the number of incorrect predictions correctly mistrusted, which is called \textit{True mistrust} ($Tm$); 
    \item the number of incorrect predictions incorrectly trusted, which is called \textit{False trust} ($Ft$); 
    \item the number of correct predictions incorrectly mistrusted, which is called \textit{False mistrust} ($Fm$).
\end{itemize}

In user evaluations, the users answer if they trust a prediction or not, making the output binary even for multi-class problems.

Consequently, \textbf{Misuse} could be defined as when the user have a high number of false trust in the predictions. Misuse is, in other words, a situation where the user performance show a low accuracy of trust. Looking at the confusion matrix in Fig.~\ref{fig:trust_conf_matr}, misuse could also be defined as a low proportion of true trust $Tt$ in the total number of trusted predictions, $Tt+Ft$. Going back to the definition of precision as the proportion of the positive predictions that are correct, we see that the definition of misuse is when the user performance has a low level of user precision, $U_{pr}$, calculated as:
\begin{equation*}
    U_{pr} = \frac{Tt}{Tt+Ft}
\end{equation*}

\textbf{Disuse} is another aspect that should be avoided. The metric could be defined as when the user performance has a high number of false mistrust, $Fm$, in relation to the number of correctly classified predictions, $Tt+Fm$. In the worst case, the user would categorise all predictions as incorrectly classified. Looking at the confusion matrix in Fig.~\ref{fig:trust_conf_matr}, a high number of false mistrust, $Fm$, would result in a low proportion of true trust, $Tt$. Disuse could in that way be said to have a low proportion of true trust in relation to the number of correctly classified predictions. Going back to the definition of recall earlier in the paper, disuse could in this way be defined as when the user has a low level of user recall, $U_{rc}$, calculated as:
\begin{equation*}
    U_{rc} = \frac{Tt}{Tt+Fm}
\end{equation*}

When defining appropriate trust, it was stated that the metric should have a high proportion of true trust, $Tt$, and true mistrust, $Tm$. At the same time, it was stated that the user should have low misuse and disuse. %, see Fig.~\ref{fig:AT_conf_matrix}. 
Misuse where defined as a low level of user precision, and disuse as a low level of user recall. In other words, when the explanations result in a high level of user performance, the $U_{pr}$ and the $U_{rc}$ should be high. In this sense, the method for measuring the level of appropriate user trust is similar to how performance of a classifier is measured with the $F_1$~score. Going back to the definition of the $F_1$~score, the metric only gives a high value to a predictor when both recall and precision are high. Since the goal of appropriate trust is to have low misuse and disuse, it is identical to the $F_1$~score. It is consequently possible to define appropriate user trust, $U_{at}$, as:
\begin{equation*}
    U_{at} = 2 * \frac{U_{pr} * U_{rc}}{U_{pr} + U_{rc}}
\end{equation*}

\section{Evaluating appropriate trust in Regression}\label{uncert}
In regression, the predictions are normally given as a point approximation, and it could therefore be challenging to evaluate the level of appropriate trust. By adopting the ideas from conformal regression, letting the user provide an interval believed to include the true target, it becomes possible to translate the point approximation to a binary situation with the true value either being inside or outside the proposed interval.

When only considering the two cases of the true target being inside or outside the interval, respectively, the same approach as described above can be used to evaluate user precision, $U_{pr}$, user recall, $U_{rc}$, and appropriate user trust, $U_{at}$.
\begin{table}[hbt!]
    \centering  
    \caption{Similarities between model and user performance metrics.}
    \begin{tabular}{
                    >{\raggedright\arraybackslash}p{0.3\linewidth}
                    >{\raggedright\arraybackslash}p{0.3\linewidth}
                    }
    
    \textbf{Model performance}  &   \textbf{User performance}\\ 
    \hline
    \addlinespace[3pt]
    high precision   &    low misuse\\
    high recall      &   low disuse\\
    $F_1$ score      &   appropriate trust\\ 
    \end{tabular}
    \label{tab:mod_at}
    \normalsize
    \hfill
\end{table}
Even if it might be important to distinguish between different kinds of errors, e.g., where the true target is above (or below) the interval suggested by the user, this does not effectively alter the proposed calculations, as user precision, $U_{pr}$, user recall, $U_{rc}$, and appropriate user trust, $U_{at}$, only has the $T_t$ as nominator. Consequently, the evaluation of different kinds of errors will not affect the proposed evaluation, but can still be evaluated by looking into the different mistakes separately. 

There are different methods for estimating the uncertainty of a regression problem. In Conformal regression \cite{johansson2014regression}, the generated confidence intervals can be adapted based on a difficulty estimation. When this is done, the interval's width indicates the uncertainty level, where a wider interval signifies a higher uncertainty and a more narrow interval represents a higher certainty. The user's level of uncertainty could in a similar way be incorporated into the evaluation by allowing them to adjust the width of the interval based on their uncertainty in the prediction, just as in Conformal regression. Thus, when the user is more certain, the width of the proposed interval could be reduced, and vice versa for situations where the user is less certain. %Suppose the given interval is within the predicted interval from the underlying model or captures the approximated value. In that case, the user has a True trust in the prediction and can be used with the confusion matrix as in Fig.~\ref{fig:trust_conf_matr} together with the metrics mentioned in the paper.

There are several different aspects of analysing user performance for regression. However, this paper highlights the possibilities of allowing capturing user uncertainty and a more user-friendly approach to evaluating appropriate user trust in regression.

\subsection{Calibrating user uncertainty in classification}
The paper has introduced a new way of defining appropriate user trust and discussed how it can be applied to the evaluation of explanations for both classification and regression. For regression, the handling of user uncertainty in the evaluations was also discussed. For classification, there is no obvious way of incorporating uncertainty in the proposed definition of appropriate user trust. However, one possible way could be to look into how uncertainty is incorporated into the probability estimates of Venn-Abers. By allowing the users to estimate their certainty in a predicted class, it might be possible to complement the proposed metrics of user precision, $U_{pr}$, user recall, $U_{rc}$, and appropriate user trust, $U_{at}$, with a method for measuring how well calibrated the users uncertainty is. Exactly how this should be done is, however, left as future work. 

\section{Example from Real World Data} \label{User_ev}
A very limited qualitative user evaluation from \cite{lofstrom2018interpretable} with six expert users is used to exemplify how the metrics can be used with real-world data. Each respondent were shown six predictions from a text classification based on a newspaper data set. The instances were divided into three categories from two classes; correctly predicted with high probability estimates, incorrect with high probability estimates, and highly uncertain predictions with a probability estimate around $0.5$. 
\begin{table}[hbt!]
    \centering  
    \caption{Answers from the respondents in the evaluation from \cite{lofstrom2018interpretable}.}
    %\footnotesize
    \begin{tabular}{
                    >{\raggedright\arraybackslash}p{0.15\linewidth}
                    >{\raggedright\arraybackslash}p{0.15\linewidth}
                    >{\raggedright\arraybackslash}p{0.15\linewidth}
                    }
    
        & \textbf{correct}  &   \textbf{incorrect}\\
    \hline
    \addlinespace[3pt]
    trust   &    11  & 7\\
    mistrust  &   1 &  5\\
    \end{tabular}
    \label{tab:mod_at}
    \normalsize
    \hfill
\end{table}

Since the highly uncertain predictions were not labelled as belonging to either class, they were excluded from the example in this paper. Therefore, the data collected from the paper include the answers from the respondents to either the highly correct or highly incorrect predictions, resulting in 12 predictions of each class. For a detailed description of the method, see \cite{lofstrom2018interpretable}

User precision is defined as the number of true trust divided with the total amount of trusted predictions, $Tf+Ft$. From table~\ref{tab:mod_at} we get that $Tt=11$ and that $Ft=7$ which gives:
\begin{equation*}
    U_{pr} = \frac{11}{11+7} = 0.61
\end{equation*}

The user precision tells us that only 61\% of the predictions trusted as correct actually were correct predictions. The respondents had, in other words, a tendency to overtrust.

In the next step we calculate the user recall to see the proportion of the correct predictions that the respondents succeeded to identify; $Tt=11$ and $Fm = 1$:
\begin{equation*}
    U_{rc} = \frac{11}{11+1} = 0.92
\end{equation*}

The result of $0.92$ in user recall shows that the users had a low level of disuse which is not surprising considering the low user precision. The metrics reveal, in other words, different aspects of the explanations, e.g., that they did not result in a low level of overtrust (or user precision). Although, it is at an acceptable level above 0.5. In \cite{lofstrom2018interpretable}, the authors write that the qualitative analysis of the answers from the respondents revealed that some of them were persuaded by the explanations to accept the predictions from the system, creating misuse. Using the suggested metrics of user precision and recall in this paper could have supported and further nuanced the analysis.

One benefit of the metric $F_1$~score is that it results in a single quality value for the entire model, making it possible to compare the results between different models. When identifying appropriate trust as the user's $F_1$~score, the results from different evaluations can be compared, circumventing the impossibility of comparing the results of subjective measurements. In the last step, we calculate the level of appropriate trust with the user $F_1$~score to get a degree of quality, possible to compare with other evaluations: 
\begin{equation*}
    U_{at} = 2*\frac{0.61*0.92}{0.61+0.92} = 0.73
\end{equation*}

Although the degree of appropriate trust is not high, it still has an acceptable level of 73\%, thanks to the relatively high level of recall. It is worth noticing that there is a trade-off between precision and recall: when precision is high, recall tends to be low, and vice versa.

%The example shows that the apparent similarities between model performance and user performance, as seen in table~\ref{tab:mod_at}, could be helpful when analysing evaluations of user performance in explanations. However, it is also crucial to remember that in explanation methods, it is often the change of appropriate trust when the users get the predictions explained that is interesting to measure. With the definition of appropriate trust as the user's performance, it is essential to have a starting value to see how the appropriate trust changes after the user has been introduced to an explanation. If the evaluation focuses on increasing the appropriate trust level, measuring the level before and after the user meets the explanations is necessary. In this sense, the evaluation gets a starting value to refer to the eventual changes and possibly evaluate if there are any significant changes in appropriate trust. 
%
The given example highlights the potential value of the apparent similarities between model performance and user performance, as seen in table~\ref{tab:mod_at}, when analysing evaluations of explanations. However, it is crucial to remember that evaluating explanation methods should also consider changes in appropriate trust. Appropriate trust, defined as the user's performance, serves as a crucial benchmark to observe if it varies after the user is introduced to an explanation. It is essential to have a starting value to see how the appropriate trust changes after the user have been introduced to an explanation. In this sense, the evaluation gets a starting value to refer to the eventual changes and possibly evaluate if there are any significant changes in appropriate trust. This approach enables a comparative analysis and facilitates the identification of significant alterations in the user's appropriate trust based on the introduction of explanations.

\section{Discussion}
In studies discussing appropriate trust, overtrust, or undertrust, it is common to analyse the metrics individually, as seen in previous works (e.g., \cite{merritt2015well} or \cite{yang20}). However, the proposed approach in this paper suggests a more comprehensive analysis by utilising the $F_1$ score as an indicator of the appropriate trust level. As mentioned earlier, in machine learning models, the F1 score reflects high performance only when both recall and precision are high. This implies that the user evaluation should exhibit low levels of misuse (overtrust) and disuse (undertrust) to reach a high level of appropriate trust. Instead of comparing three separate metrics, the approach considers a single metric demonstrating the relationship between overtrust and undertrust, penalising extreme values. In other words, when both overtrust and undertrust are relatively low, the level of appropriate trust is higher than in situations where only one is high.

By defining overtrust and undertrust as low levels of precision and recall, we can compare the model's performance with the user's and analyse whether the explanations effectively highlight the instances where the model struggles to make accurate predictions. For example, suppose the model exhibits low recall, indicating that it fails to correctly identify instances without mislabeling them. In that case, we aim for users to identify this issue with the help of explanations. Therefore, the evaluation should show an increased level of precision, i.e., a decrease in overtrust, indicating that users are becoming more cautious in trusting the model's predictions.

Moreover, the proposed definitions enable user evaluations to incorporate other metrics associated with model performance. In scenarios where the primary interest is to minimise the false positives, although the false negatives are still important, the $F_2$ metric can be used, focusing more on undertrust. It can, e.g., be employed to assess how well users, with the assistance of explanations, can identify this type instances. 

\section{Conclusion}
Explanation methods are essential for increasing the quality of decisions made by Decision Support Systems (DSS) users. However, evaluating these methods poses a significant challenge, involving many highly subjective criteria. 
To address the challenge, this paper proposes a new approach that treats human users as black box models and evaluates their performance using existing metrics for model performance. Specifically, the paper compares the definitions and usage of appropriate trust (or calibrated trust) in literature to the metrics for evaluating model performance and highlights their strong similarities.
By using the same metrics for both AI models and human users, the paper provides a new range of questions that can be answered, such as not only the level of mistrust and distrust the user has but also the calibration of their trust (i.e., whether they identify correct predictions more often than incorrect ones), and the relative ease of identifying accurate and inaccurate predictions. Established metrics also offer a well-defined and widely accepted framework for analysis.
To demonstrate the feasibility of this approach, the paper presents an analysis of results from a real-world user evaluation. The proposed methodology sheds light on users' performance and provides valuable insights into evaluating the strengths and weaknesses of different explanation methods.

In summary, this paper presents a novel approach to evaluating explanation methods in XAI by treating human users as black box models and using established metrics for evaluating model performance. This approach provides a clear and objective framework for analysis and has the potential to offer new insights into the performance of users and the effectiveness of different explanation methods.

%: these relate to both the level of mistrust and distrust the user has and also the calibration of their trust (i.e., whether they identify correct predictions more often than incorrect ones), and the relative ease of identifying accurate and inaccurate predictions. Established metrics also offer a well-defined and widely accepted framework for analysis.
%, such as satisfaction and expectation. Moreover, including human actors further complicates the evaluation process, as they are highly complex and challenging to measure.

\bibliographystyle{ieeetr}
\bibliography{IEEEabrv,Bibliography}

\begin{thebibliography}{10}

\bibitem{zhou2021evaluating}
J.~Zhou, A.~H. Gandomi, F.~Chen, and A.~Holzinger, ``Evaluating the quality of
  machine learning explanations: A survey on methods and metrics,'' {\em
  Electronics}, vol.~10, no.~5, p.~593, 2021.

\bibitem{ribeiro2016should}
M.~T. Ribeiro, S.~Singh, and C.~Guestrin, ``Why should i trust you?" explaining
  the predictions of any classifier,'' in {\em Proceedings of the 22nd ACM
  SIGKDD international conference on knowledge discovery and data mining},
  pp.~1135--1144, 2016.

\bibitem{AIHLEG}
{The European Commission Independent High-Level Expert Group on Artificial
  Intelligence}, ``Ethics {G}uidelines for {T}rustworthy {AI},'' 2019.

\bibitem{DavidGunning2017}
{David Gunning}, ``{Explainable Artificial Intelligence}.'' Web, 2017.
\newblock {DARPA}.

\bibitem{dimanov2020you}
B.~Dimanov, U.~Bhatt, M.~Jamnik, and A.~Weller, ``You shouldn’t trust me:
  Learning models which conceal unfairness from multiple explanation
  methods.,'' {\em Frontiers in Artificial Intelligence and Applications: ECAI
  2020}, 2020.

\bibitem{Lofstrom2022}
H.~L{\"o}fstr{\"o}m, K.~Hammar, and U.~Johansson, ``A meta survey of quality
  evaluation criteria in explanation methods,'' in {\em Intelligent Information
  Systems} (J.~De~Weerdt and A.~Polyvyanyy, eds.), (Cham), pp.~55--63, Springer
  International Publishing, 2022.

\bibitem{hoffman2018metrics}
R.~R. Hoffman, S.~T. Mueller, G.~Klein, and J.~Litman, ``Metrics for
  explainable ai: Challenges and prospects,'' {\em arXiv preprint
  arXiv:1812.04608}, 2018.

\bibitem{Guidotti18}
R.~{Guidotti}, A.~{Monreale}, S.~{Ruggieri}, F.~{Turini}, D.~{Pedreschi}, and
  F.~{Giannotti}, ``{A Survey Of Methods For Explaining Black Box Models},''
  {\em ArXiv e-prints}, Feb. 2018.

\bibitem{molnar2020interpretable}
C.~Molnar, {\em Interpretable Machine Learning}.
\newblock Leanpub, 2~ed., 2022.

\bibitem{moradi2021post}
M.~Moradi and M.~Samwald, ``Post-hoc explanation of black-box classifiers using
  confident itemsets,'' {\em Expert Systems with Applications}, vol.~165,
  p.~113941, 2021.

\bibitem{holzinger2019causability}
A.~Holzinger, G.~Langs, H.~Denk, K.~Zatloukal, and H.~M{\"u}ller, ``Causability
  and explainability of artificial intelligence in medicine,'' {\em Wiley
  Interdisciplinary Reviews: Data Mining and Knowledge Discovery}, vol.~9,
  no.~4, p.~e1312, 2019.

\bibitem{doshi2017towards}
F.~Doshi-Velez and B.~Kim, ``Towards a rigorous science of interpretable
  machine learning,'' {\em arXiv preprint arXiv:1702.08608}, 2017.

\bibitem{gunning19}
D.~Gunning and D.~W. Aha, ``Darpa’s explainable artificial intelligence
  program,'' {\em AI Magazine}, vol.~40, no.~2, pp.~44--58, 2019.

\bibitem{Carvalho19}
D.~V. Carvalho, E.~M. Pereira, and J.~S. Cardoso, ``Machine learning
  interpretability: A survey on methods and metrics,'' {\em Electronics},
  vol.~8, p.~832, 2019.

\bibitem{hoff2015trust}
K.~A. Hoff and M.~Bashir, ``Trust in automation: Integrating empirical evidence
  on factors that influence trust,'' {\em Human factors}, vol.~57, no.~3,
  pp.~407--434, 2015.

\bibitem{jacovi2021formalizing}
A.~Jacovi, A.~Marasovi{\'c}, T.~Miller, and Y.~Goldberg, ``Formalizing trust in
  artificial intelligence: Prerequisites, causes and goals of human trust in
  ai,'' in {\em Proceedings of the 2021 ACM conference on fairness,
  accountability, and transparency}, pp.~624--635, 2021.

\bibitem{yang20}
F.~Yang, Z.~Huang, J.~Scholtz, and D.~L. Arendt, ``How do visual explanations
  foster end users' appropriate trust in machine learning?,'' in {\em
  Proceedings of the 25th International Conference on Intelligent User
  Interfaces}, pp.~189--201, 2020.

\bibitem{sheridan2002humans}
T.~B. Sheridan, T.~B. Sheridan, K.~Maschinenbauingenieur, T.~B. Sheridan, and
  T.~B. Sheridan, {\em Humans and automation: System design and research
  issues}, vol.~280.
\newblock Human Factors and Ergonomics Society Santa Monica, CA, 2002.

\bibitem{adya2020stressed}
M.~Adya and G.~Phillips-Wren, ``Stressed decision makers and use of decision
  aids: a literature review and conceptual model,'' {\em Information Technology
  \& People}, vol.~33, no.~2, pp.~710--754, 2020.

\bibitem{alvarado2014reliance}
J.~A. Alvarado-Valencia and L.~H. Barrero, ``Reliance, trust and heuristics in
  judgmental forecasting,'' {\em Computers in human behavior}, vol.~36,
  pp.~102--113, 2014.

\bibitem{buccinca2020proxy}
Z.~Bu{\c{c}}inca, P.~Lin, K.~Z. Gajos, and E.~L. Glassman, ``Proxy tasks and
  subjective measures can be misleading in evaluating explainable ai systems,''
  in {\em Proceedings of the 25th international conference on intelligent user
  interfaces}, pp.~454--464, 2020.

\bibitem{geron2022hands}
A.~G{\'e}ron, {\em Hands-on machine learning with Scikit-Learn, Keras, and
  TensorFlow}.
\newblock " O'Reilly Media, Inc.", 2022.

\bibitem{vovk2005algorithmic}
V.~Vovk, A.~Gammerman, and G.~Shafer, {\em Algorithmic learning in a random
  world}, vol.~29.
\newblock Springer, 2005.

\bibitem{johansson2014regression}
U.~Johansson, H.~Bostr{\"o}m, T.~L{\"o}fstr{\"o}m, and H.~Linusson,
  ``Regression conformal prediction with random forests,'' {\em Machine
  learning}, vol.~97, pp.~155--176, 2014.

\bibitem{Lambrou2015}
A.~Lambrou, I.~Nouretdinov, and H.~Papadopoulos, ``Inductive venn prediction,''
  {\em Annals of Mathematics and Artificial Intelligence}, vol.~74, no.~1,
  pp.~181--201, 2015.

\bibitem{mcdermott2019practical}
P.~L. McDermott and R.~N.~t. Brink, ``Practical guidance for evaluating
  calibrated trust,'' in {\em Proceedings of the Human Factors and Ergonomics
  Society Annual Meeting}, vol.~63, pp.~362--366, SAGE Publications Sage CA:
  Los Angeles, CA, 2019.

\bibitem{arrieta20}
A.~B. Arrieta, N.~D{\'\i}az-Rodr{\'\i}guez, J.~Del~Ser, A.~Bennetot, S.~Tabik,
  A.~Barbado, S.~Garc{\'\i}a, S.~Gil-L{\'o}pez, D.~Molina, R.~Benjamins, {\em
  et~al.}, ``Explainable artificial intelligence (xai): Concepts, taxonomies,
  opportunities and challenges toward responsible ai,'' {\em Information
  Fusion}, vol.~58, pp.~82--115, 2020.

\bibitem{chromik2020taxonomy}
M.~Chromik and M.~Schuessler, ``A taxonomy for human subject evaluation of
  black-box explanations in xai.,'' in {\em ExSS-ATEC@ IUI}, 2020.

\bibitem{das2020opportunities}
A.~Das and P.~Rad, ``Opportunities and challenges in explainable artificial
  intelligence (xai): A survey,'' {\em arXiv preprint arXiv:2006.11371}, 2020.

\bibitem{adadi2018peeking}
A.~Adadi and M.~Berrada, ``Peeking inside the black-box: A survey on
  explainable artificial intelligence (xai),'' {\em IEEE Access}, vol.~6,
  pp.~52138--52160, 2018.

\bibitem{wang2019Designing}
D.~Wang, Q.~Yang, A.~Abdul, and B.~Y. Lim, ``Designing theory-driven
  user-centric explainable ai,'' in {\em Proceedings of the 2019 CHI Conference
  on Human Factors in Computing Systems}, CHI '19, (New York, NY, USA),
  p.~1–15, Association for Computing Machinery, 2019.

\bibitem{mohseni2018multidisciplinary}
S.~Mohseni, N.~Zarei, and E.~D. Ragan, ``A multidisciplinary survey and
  framework for design and evaluation of explainable ai systems,'' {\em arXiv},
  pp.~arXiv--1811, 2018.

\bibitem{gunning2021darpa}
D.~Gunning, E.~Vorm, Y.~Wang, and M.~Turek, ``Darpa’s explainable ai (xai)
  program: A retrospective,'' {\em Authorea Preprints}, 2021.

\bibitem{zhang2018explainable}
Y.~Zhang and X.~Chen, ``Explainable recommendation: A survey and new
  perspectives,'' {\em arXiv preprint arXiv:1804.11192}, 2018.

\bibitem{merritt2015well}
S.~M. Merritt, D.~Lee, J.~L. Unnerstall, and K.~Huber, ``Are well-calibrated
  users effective users? associations between calibration of trust and
  performance on an automation-aided task,'' {\em Human Factors}, vol.~57,
  no.~1, pp.~34--47, 2015.

\bibitem{lee2004trust}
J.~D. Lee and K.~A. See, ``Trust in automation: Designing for appropriate
  reliance,'' {\em Human factors}, vol.~46, no.~1, pp.~50--80, 2004.

\bibitem{gunning2019xai}
D.~Gunning, M.~Stefik, J.~Choi, T.~Miller, S.~Stumpf, and G.-Z. Yang,
  ``Xai—explainable artificial intelligence,'' {\em Science robotics},
  vol.~4, no.~37, p.~eaay7120, 2019.

\bibitem{holzinger2020measuring}
A.~Holzinger, A.~Carrington, and H.~M{\"u}ller, ``Measuring the quality of
  explanations: the system causability scale (scs),'' {\em KI-K{\"u}nstliche
  Intelligenz}, pp.~1--6, 2020.

\bibitem{lofstrom2018interpretable}
H.~L{\"o}fstr{\"o}m, T.~L{\"o}fstr{\"o}m, and U.~Johansson, ``Interpretable
  instance-based text classification for social science research projects,''
  {\em Archives of Data Science, Series A}, vol.~5, no.~1, 2018.

\end{thebibliography}
\end{document}